%% file: graphite2014.tex
\documentclass[copyright,creativecommons,noderivs]{eptcs}
\providecommand{\event}{GRAPHITE 2007} 
\pdfoutput=1  

\usepackage{amsmath}
\usepackage{amssymb}
\usepackage{amsthm}
\newtheorem{definition}{Definition}
\usepackage{mathtools}
\usepackage{algorithm}
\usepackage{algpseudocode}
\usepackage{groove2tikz4fujaba}
\usepackage{caption}
\usepackage{subcaption}

\title{Graph Transformation Planning via Abstraction}
\author{Steffen Ziegert
  \institute{Department of Computer Science\\University of Paderborn}
  \email{steffen.ziegert@uni-paderborn.de}
}
\def\titlerunning{Graph Transformation Planning via Abstraction}
\def\authorrunning{S. Ziegert}
\begin{document}
\maketitle

\begin{abstract}
  Modern software systems increasingly incorporate self-* behavior to adapt to changes in the environment at runtime.
  Such adaptations often involve reconfiguring the software architecture of the system.
  Many systems also need to \emph{manage} their architecture themselves, i.e., they need a \emph{planning component} to autonomously decide which reconfigurations to execute to reach a desired target configuration.
  For the specification of reconfigurations, we employ graph transformations systems (GTS) due to the close relation of graphs and UML object diagrams.
  We solve the resulting planning problems with a planning system that works directly on a GTS.
  It features a domain-independent heuristic that uses the solution length of an abstraction of the original problem as an estimate.
  Finally, we provide experimental results on two different domains, which confirm that our heuristic performs better than another domain-independent heuristic which resembles heuristics employed in related work.
\end{abstract}


\input{1_introduction}

\input{2_fundamentals}

\input{3_approach}

\input{4_evaluation}

\input{5_related_work}

\input{6_conclusion}

\section*{Acknowledgement}
The author would like to thank Amir Shayan Ahmadian and Johannes Geismann for major contributions to the implementation and valuable discussions.

\bibliographystyle{eptcsalpha}
\bibliography{references}
\end{document}

%% file: 1_introduction.tex
\section{Introduction}
\label{sec:introduction}

Modern software systems increasingly incorporate self-* behavior to adapt to changes in the environment at runtime.
Such adaptations often involve reconfiguring the software architecture of the system.
However, the ability to \emph{perform} reconfiguration might not be enough.
Many systems also need to \emph{manage} their architecture themselves, i.e., they need a \emph{planning component} to autonomously decide which reconfigurations to execute to reach a desired target configuration \cite{kramer07_anArchitecturalChallenge}.

Graph transformations systems (GTS) \cite{ehrig06_fundamentals} have been deemed a suitable formalism for the specification of reconfigurations due to their close relation to graphs and UML object diagrams \cite{leMetayer98_describingSoftwareArchitectureStyles,wermelinger02_aGraphTransformationApproach}.
They have been used for the verification of reconfiguration operations in various approaches, e.g., \cite{becker06_symbolicInvariantVerification, rensink08_explicit}.
Planning of architecture reconfiguration has also been covered before, e.g., in \cite{estler11_heuristic} for coordinating behavior in cyber-physical systems and in \cite{tichy11_planningSelfAdaptation} for planning a self-healing process in automotive systems.
The involved planning problems are usually solved by one the following two approaches: either a translation into a dedicated planning language, i.e., the Planning Domain Definition Language (PDDL) \cite{fox03_pddl21}, is performed or a planning system is developed that works directly on a GTS.

Both approaches have their drawbacks.
Translation-based approaches suffer from a different expressiveness of GTSs and PDDL: while the creation and deletion of objects is a fundamental feature of GTSs, there is no such thing in PDDL.
By not allowing to de-/instantiate objects, PDDL maintains a finite state space.
To handle de-/instantiation of objects nevertheless, a modeling workaround can be used that declares all uninstantiated objects in the initial state, but uses a predicate to state their actual existence, as in \cite{tichy11_planningSelfAdaptation,ziegert13_temporalReconfigurationPlans}.
However, the workaround is based on the assumption that a maximal number of objects is known beforehand or can be deduced from the GTS.

Planning systems for GTSs on the other hand are not highly evolved.
Up until today, there are only few systems that use domain-independent heuristics to guide their search through the state space of a GTS.
They employ simple domain-independent heuristics that compute values for the structural similarity of the current configuration and the target configuration.
Multiple such \emph{similarity-based} heuristic functions are presented in \cite{edelkamp06_heuristicSearch}.
Several of them are different variants of counting the nodes and edges that have to be created or deleted to reach the target configuration, e.g., they differ in whether or not it is allowed to rely on the identity of nodes and edges, and then using this number as a distance measure.
Another slightly different variant of such a heuristic has been presented in \cite{snippe11_usingHeuristicSearch}.

In this paper, we propose a new GTS-based planning system.
It employs a domain-independent heuristic function that can be used in different search algorithms.
The heuristic function computes the solution length of an abstraction of the original problem as an estimate for a given state.
The abstract problem reinterprets certain parts of the rules' application conditions and is thus easier to solve than the original problem.
As part of our contribution, we compare the performance of our heuristic against the performance of a similarity-based heuristic.

The next section introduces graph transformations systems and the notion of planning problems on graph transformations systems.
In Section~\ref{sec:example} we present an application example, which is used to explain our heuristic approach in Section~\ref{sec:approach}.
The evaluation of our approach is given in Section~\ref{sec:evaluation}.
We discuss related work in Section~\ref{sec:related_work} before concluding in Section~\ref{sec:conclusion}.

%% file: 2_fundamentals.tex
\section{Planning with Graph Transformations}
\label{sec:fundamentals}

A graph transformation system (GTS) consists of a set of graph transformation rules and an initial graph.
The graph transformation rules can be applied to the initial graph and its resulting successor graphs to construct the state space of the GTS.
The underlying theory of GTS is based on graphs and graph morphisms.

\begin{definition}[Graph, Graph Morphism]
\label{def:morphism}
A \emph{graph} $G = (V_G, E_G, src_G, tgt_G)$ consists of a set of nodes $V_G$, a set of edges $E_G$, and source and target functions $src_G, tgt_G: E_G \rightarrow V_G$.
A \emph{graph morphism} $f: G \rightarrow H$ between two graphs is a pair of mappings $f = (f_E, f_V)$ with $f_E: E_G \rightarrow E_H$ and $f_V: V_G \rightarrow V_H$ such that $f_V \circ src_G = src_H \circ f_E$ and $f_V \circ tgt_G = tgt_H \circ f_E$.
A graph morphism $f = (f_E, f_V)$ is \emph{injective} if $f_E$ and $f_V$ are injective.
\end{definition}

A graph morphism is a mapping of nodes and edges of one graph to nodes and edges of another graph such that the source and target nodes of edges are preserved.
Such morphisms are used in graph transformation rules to define which nodes and edges are created, deleted, or preserved when the rule is applied to a graph.

\begin{definition}[Graph Transformation Rule]
\label{def:rule}
A \emph{graph transformation rule} $p = (L, R, r)$ consists of two graphs $L$ and $R$, called \emph{left-hand side (LHS)} and \emph{right-hand side (RHS)}, and an injective partial graph morphism $r: L \rightarrow R$, called \emph{rule morphism}.
Given a graph transformation rule $p$ and a match $m: L \rightarrow G$ of its LHS into a \emph{host graph} $G$, the \emph{direct derivation} from $G$ with $p$ at $m$, written $G \xRightarrow{p,m} H$, is the pushout of $r$ and $m$ in $Graph^P$, the category of graphs and partial graph morphisms, as shown below. \cite{ehrig97_handbookChapterSPO}
\begin{center}
  \input{figures/fig_pushout.tex}
\end{center}
\end{definition}

Whether a graph transformation rule can be applied to a graph depends on whether a match of its LHS to the graph can be found.
Multiple such matches result in multiple direct derivations and thus in multiple successor graphs.
When a rule is applied that specifies the deletion of a node, dangling edges might occur.
The above definition follows the single pushout approach (SPO), which results in dangling edges being deleted as well.
Furthermore, we employ injective matchings: each node (and edge) of the rule maps to a distinct node (or edge) of the graph.

To restrict the applicability of a rule, negative application conditions (NAC) can be used that forbid specific graph structures from being present in the graph.

\begin{definition}[Negative Application Condition]
Let $p = (L, R, r)$ be a graph transformation rule, $G$ a graph, and $m: L \rightarrow G$ a match.
A \emph{negative application condition (NAC)} is a tuple $NAC = (N, n)$ with $n: L \rightarrow N$ and $n$ being injective.
If $\neg \exists q: N \rightarrow G$ such that $q \circ n = m$, then $m$ \emph{satisfies} $NAC$, written $m \models NAC$.
\end{definition}

We also write $p = (L, R, r, \mathcal{N})$, where $\mathcal{N}$ is a set of NACs, when we want to explicitly refer to the NACs of a graph transformation rule $p$.
Now we have everything we need for the definition of graph transformation systems.

\begin{definition}[Graph Transformation System]
A \emph{graph transformation system} $S = (\mathcal{R}, G_0)$ consists of a set of graph transformation rules $\mathcal{R}$ and an initial graph $G_0$.
\end{definition}

A planning problem on a graph transformation system also involves a graph pattern, which defines valid target configurations.
The specification of graph patterns also support NACs.

\begin{definition}[Graph Pattern]
A \emph{graph pattern} $P = (L, \mathcal{N})$ consists of a graph $L$ and a set of NACs $\mathcal{N}$ where each $NAC \in \mathcal{N}$ is a tuple $NAC = (N, n)$ with $n: L \rightarrow N$ and $n$ being injective.
\end{definition}

Having a means of specifying valid target configurations of a planning problem, we can now define the planning problem on a graph transformation system.

\begin{definition}[Planning Problem]
A \emph{graph transformation planning problem} $\mathcal{P} = (\mathcal{R}, G_0, P_{tgt})$ consists of a set of graph transformation rules $\mathcal{R}$, an initial graph $G_0$, and a target graph pattern $P_{tgt} = (L_{tgt}, \mathcal{N}_{tgt})$.
A \emph{plan} for $\mathcal{P}$ is a sequence of direct derivations $G_0 \Rightarrow \ldots \Rightarrow G_k$ such that the target graph pattern $P_{tgt}$ has a match in $G_k$.
\end{definition}

%% file: figures/fig_pushout.tex
\begin{tikzpicture}[y=0.7cm,x=1.2cm,descr/.style={fill=white,inner sep=2.5pt}]
\node (A) at (-1,1) {$L$};
\node (B) at (1,1) {$R$};
\node (C) at (-1,-1) {$G$};
\node (D) at (1,-1) {$H$};
\node (E) at (0,0) {$(PO)$};
\draw [->] (A) edge node[descr] {$r$} (B);
\draw [->] (C) edge node[descr] {$r'$} (D);
\draw [->] (A) edge node[descr] {$m$} (C);
\draw [->] (B) edge node[descr] {$m'$} (D);
\end{tikzpicture}

%% file: 3_approach.tex
\section{Application Example}
\label{sec:example}

We consider the reconfiguration of Electronic Control Units (ECUs) in automotive systems as an application scenario.
In current development, software components are deployed on ECUs at design-time.
The AUTOSAR consortium proposed a component-based software architecture standard\footnote{AUTOSAR specifications are available at \url{http://www.autosar.org/index.php?p=3}.} for the development of ECU software.
Following the AUTOSAR standard, a Runtime Environment (RTE) is generated out of a predefined set of components.
The RTE acts as a middleware that connects the software components with Basic Software (BSW) that controls the hardware.
We expect that in the future software components can be deployed at ECUs at runtime.
This allows to react to hardware failures or adapt to low levels of energy by runtime reconfiguration.
Such runtime reconfigurations are executed according to reconfiguration plans, which are computed in soft real-time.
In the context of our application example, a planning task might be to shut down an ECU due to a recognized hardware failure or reboot an ECU due to a re-deployment of the RTE middleware after a software upgrade.

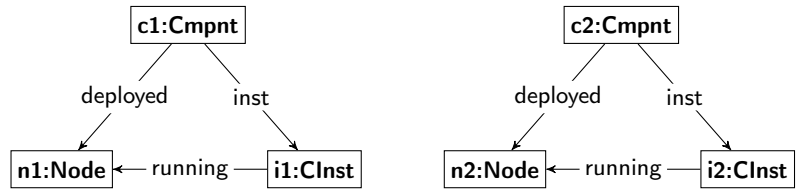
\begin{figure}[ht]
  \centering
  \input{figures/ECUs-export/start.tikz}
  \caption{Initial configuration}
  \label{fig:start}
\end{figure}

Consider the graph in Figure~\ref{fig:start} as the initial state for the planning problem.
There are two ECUs, \texttt{n1} and \texttt{n2}, and two software components, \texttt{c1} and \texttt{c2}.
For each component there is an instance that is running on one of the ECUs.

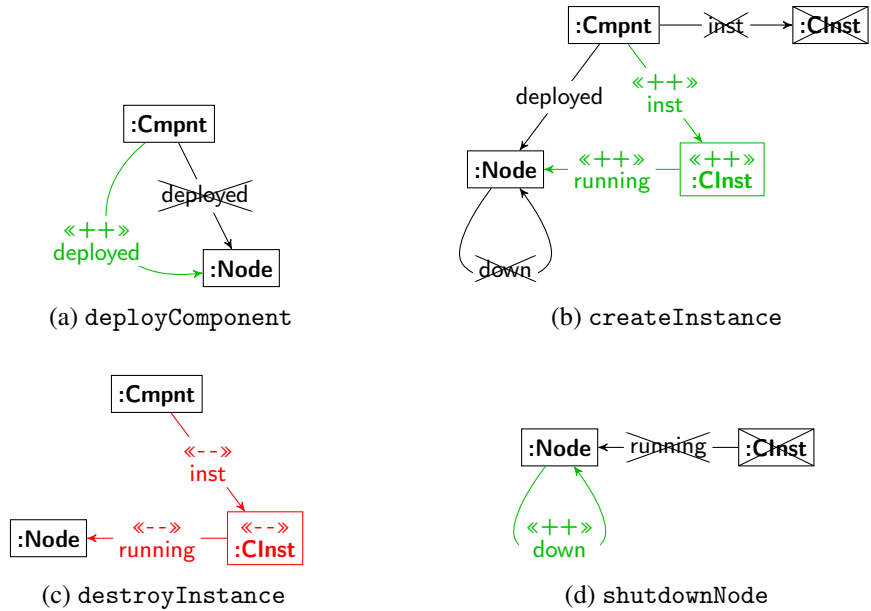
\begin{figure}[ht]
  \centering
  \begin{subfigure}[b]{0.4\textwidth}
	\centering
    \input{figures/ECUs-export/deploy.tikz}
    \caption{\texttt{deployComponent}}
    \label{fig:deployComponent}
    \vspace*{5mm}
  \end{subfigure}%
  ~ 
  \begin{subfigure}[b]{0.4\textwidth}
	\centering
    \input{figures/ECUs-export/createInst.tikz}
    \caption{\texttt{createInstance}}
    \label{fig:createInst}
    \vspace*{5mm}
  \end{subfigure}
  ~ 
  \begin{subfigure}[b]{0.4\textwidth}
	\centering
    \input{figures/ECUs-export/destroyInst.tikz}
    \caption{\texttt{destroyInstance}}
    \label{fig:destroyInst}
    \vspace*{5mm}
  \end{subfigure}
  ~ 
  \begin{subfigure}[b]{0.4\textwidth}
	\centering
    \input{figures/ECUs-export/shutdown.tikz}
    \caption{\texttt{shutdownNode}}
    \label{fig:shutdownNode}
    \vspace*{5mm}
  \end{subfigure}
  \vspace*{-5mm}
  \caption{Graph transformation rules}
  \label{fig:rules}
\end{figure}

Figures~\ref{fig:deployComponent} to~\ref{fig:shutdownNode} show the rules that are relevant for our application scenario.
We use the stereotype \texttt{<<++>>} to denote elements that are only available in the RHS and \texttt{<<-\:\!->>} to denote elements that are only available in the LHS.
Crossed out elements denote NACs.
If crossed out elements are adjacent to each other, they belong to the same NAC.
In Figure~\ref{fig:deployComponent} component data is deployed on an ECU such that the component can be instantiated.
Figure~\ref{fig:createInst} specifies the creation of a component instance.
In Figure~\ref{fig:destroyInst} an instance that is running on an ECU is destroyed.
Figure~\ref{fig:shutdownNode} specifies shutting down an ECU if no instances are running on it.

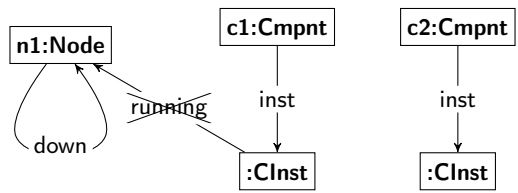
\begin{figure}[ht]
  \centering
  \input{figures/ECUs-export/goal.tikz}
  \caption{Target graph pattern}
  \label{fig:goal}
\end{figure}

The target is specified as a graph pattern in Figure~\ref{fig:goal}.
It states that ECU \texttt{n1} should be shut down and components \texttt{c1} and \texttt{c2} should both be instantiated.
Since a component instance of \texttt{c1} is running on \texttt{n1} in the initial state, we added a NAC disallowing component instances of \texttt{c1} to run on \texttt{n1} in goal states.

A sample plan arriving in a valid target configuration deploys component \texttt{c1} at ECU \texttt{n2} in the first step, then destroys the component instance \texttt{i1}, then shuts down \texttt{n1}, and at last creates a new instance of \texttt{c1} that runs on \texttt{n2}.
To compute such plans, our planning system employs a heuristic function that solves for each encountered state in its state space an abstraction of the original problem.

\section{Solving the Planning Problem via Abstraction}
\label{sec:approach}

Our technique to solve the GT planning problem is an informed search in the state space of the GTS.
The main feature of the system is the heuristic function that tells the search algorithm, e.g., A*, Best-First (BF), or Enforced Hill-Climbing (EHC), which state to expand next.
The heuristic function itself solves a planning problem that is an abstraction of the original problem considered from a given state.
Information gained during solving the abstract problem is used to guide the search of the original problem.

\subsection{Abstract State Sequences}
\label{subsec:heuristic}

The abstraction used by the heuristic function combines two ideas.
The first idea is to \emph{relax} the applicability of a rule, i.e., to apply only changes that enable subsequent rule applications, but discard changes that disable subsequent rule applications.
In essence, this is realized by not deleting elements when rules are applied (to relax LHS matching) and the use of labels, which allow to reinterpret (and thus relax) NAC matching.
The second idea is to apply all applicable transformations \emph{in parallel} to construct the next (abstract) state, instead of choosing one state to expand next.
This results in a linear (abstract) state space and thus allows to solve the abstract problem efficiently.
Two applicable transformations cannot be in conflict with each other due to the applied relaxation.

\begin{figure}[t]
  \centering
  \includegraphics[width=\textwidth]{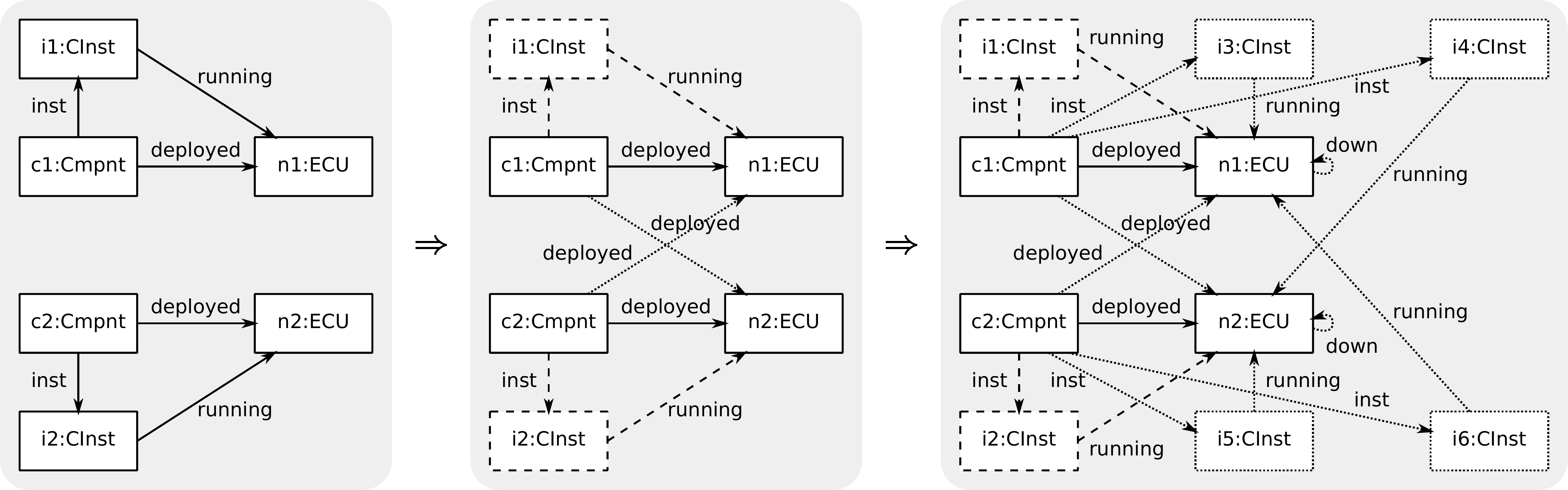}
  \caption{An abstract state sequence}
  \label{fig:parallelRelaxedPlan}
\end{figure}

Figure~\ref{fig:parallelRelaxedPlan} shows the abstract state sequence from the initial state of the problem to the first state that satisfies the target graph pattern.
Each transition corresponds to one parallel and relaxed application of all applicable transformations.
In the first transition, all reconfigurations deploying components are executed in parallel, as well as all reconfigurations destroying component instances.
In the second transition, new instances are created and both ECUs are shutdown.

The relaxation used during the construction of the abstract state sequence is supposed to discard the deletion of elements and reinterpret certain parts of the rules' application conditions.
Therefore, each element that is supposed to be deleted according to the rule morphism of an applicable transformation, is maintained in the successor graph $G_{succ}$, but is labeled with \emph{\textsf{deleted}} instead.
Each element that is supposed to be created, is added to $G_{succ}$ as usual and labeled with \emph{\textsf{created}}.
Elements labeled with \emph{\textsf{deleted}} and \emph{\textsf{created}} correspond to the dashed and dotted elements of Figure~\ref{fig:parallelRelaxedPlan}, respectively.
Labeling these elements gives the option to disregard them when considering the applicability of transformations in later iterations.
This has happened in the second transition of Figure~\ref{fig:parallelRelaxedPlan} with the component instances \texttt{i1} and \texttt{i2}: since they have been labeled as \emph{\textsf{deleted}} by the first transition, new instances can be created during the second transition by applying the \texttt{createInstance} rule.
In general terms, in order for a transformation to be considered applicable (despite a matching NAC), there has to be at least one element that is labeled with \emph{\textsf{deleted}} or \emph{\textsf{created}} in the part of the host graph that is matched by the NAC.
If there are multiple NACs or a NAC has multiple matches, then each NAC match has to contain at least one element that is labeled with \emph{\textsf{deleted}} or \emph{\textsf{created}} for the transformation to be applicable.

The generation of the abstract state sequence stops as soon as the abstract planning problem is solved, i.e., the abstract state sequence reached a state that satisfies the target graph pattern.
However, it is also possible that there is no path from the initial abstract state to a state that satisfies the target graph pattern.
Therefore, we also abort the generation of the abstract state sequence after we generated $x$ successor states, where $x$ is two times the heuristic value of the initial state of the concrete planning problem.

\subsection{Heuristic Values}
\label{subsec:revision}

Our planning system performs a state space exploration by successively choosing a state and expanding it, i.e., applying each rule at each possible match to generate its successor states.
To decide which state to expand next, the system calculates a heuristic value for each unexpanded state.
Calculating a heuristic value for a state involves generating the abstract state sequence starting in this state until we reach a state that satisfies the target graph pattern.

Having constructed the abstract state sequence, a naive idea would be to use its length as heuristic estimate.
Although the abstract state sequence is expected to be shorter for states which are near to a goal state and longer for states which are further away from a goal state, this value is still rather imprecise.
A better idea is to give the approximate number of \emph{individual} reconfigurations needed for reaching the goal state.
However, we cannot simply count all applied reconfigurations per transition to calculate this number, because this would include a lot of reconfigurations that were \emph{not} needed to reach the goal state.
The reconfigurations that \emph{were} needed to reach the goal state are called a \emph{relaxed plan} and their number is called the \emph{length} of the relaxed plan.

Our approach to calculate this number incorporates rule application information into the newly created elements of each successor graph.
Each created element is labeled with information about the transformation that caused its creation.
This label consists of the iteration number of the successor graph creation loop, the name of the applied rule, and a distinct identifier for the match of the rule to the host graph.
As an example, the \texttt{deployed} edge from component \texttt{c1} to ECU \texttt{n2} in Figure~\ref{fig:parallelRelaxedPlan} is labeled with \emph{\textsf{(iteration \#1, '\texttt{deployComponent}', match \#1)}}.

When the goal match is found, we can count the number of distinct rule application labels that are contained in the elements of the goal match.
This number is the number of transformations needed to create the elements in the goal match.
However, these labels contains only labels of elements that appear \emph{directly} in the goal match.
It does not yet contain labels of elements that were needed to \emph{arrive} at the goal match.
An example for this is the label of the aforementioned \texttt{deployed} edge from component \texttt{c1} to ECU \texttt{n2}.
While the \texttt{deployed} edge is not contained in the goal match, its creation during the first transition was necessary for the application of \emph{another} transformation during the second transition to create an element in the goal match.
In this example, the application of \texttt{createInstance} that creates component instance \texttt{i4} during the second transition required the \texttt{deployed} edge from \texttt{c1} to \texttt{n2}.
Our approach includes labels of such elements, i.e., elements that were needed to arrive at the goal match, when counting the rule application labels in the goal match: each element created by a transformation---in addition to its own label---inherits the labels of all elements in the LHS match of the rule application.
In the example of Figure~\ref{fig:parallelRelaxedPlan}, the rule application label of the \texttt{deployed} edge created during the first transition is propagated to the newly created instance \texttt{i4} during the second transition.

Elements that have been marked as \emph{\textsf{deleted}} are handled similarly.
For example, the component instance \texttt{i1} receives the rule application label \emph{\textsf{(iteration \#1, '\texttt{destroyInstance}', match \#1)}} when it is marked as \emph{\textsf{deleted}} by the application of the \texttt{destroyInstance} rule.
Labels of elements being marked as \emph{\textsf{deleted}} are propagated to newly created (or deleted) elements if the labeled element is contained in a NAC match, e.g., the label of \texttt{i1} is propagated to the \texttt{down} edge at ECU \texttt{n1} when \texttt{shutdownNode} is applied during the second transition.
Note, that elements being marked as \emph{\textsf{deleted}} inherit labels in the same manner as elements marked as \emph{\textsf{created}}: they inherit labels of created elements if contained in the LHS match and labels of deleted elements if contained in the NAC match.
By inheriting the labels of other elements, the elements in the goal match do not only contain labels of transformations that directly created them, but also about all prior transformations that made their creation possible (whether by means of element creation or deletion).

The heuristic value is now simply defined as the number of labels that are attached to all elements in the goal match.
If there are multiple goal matches, we use the smaller value.
Applied to the example of Figure~\ref{fig:parallelRelaxedPlan}, the goal match containing \texttt{i4} and \texttt{i2} would result in a heuristic value of 4.

%% file: figures/ECUs-export/start.tikz
%
\begin{tikzpicture}[
scale=\tikzscale*1.2]
\node[node] (n0)  at (1.200, -0.100) {\ml{\textbf{c1:Cmpnt}}};
\node[node] (n1)  at (0.500, -0.900) {\ml{\textbf{n1:Node}}};
\node[node] (n2)  at (1.900, -0.900) {\ml{\textbf{i1:CInst}}};
\node[node] (n3)  at (3.600, -0.100) {\ml{\textbf{c2:Cmpnt}}};
\node[node] (n4)  at (2.900, -0.900) {\ml{\textbf{n2:Node}}};
\node[node] (n5)  at (4.300, -0.900) {\ml{\textbf{i2:CInst}}};
\path[edge] (n0)  -- node[lab]{deployed} (n1) ;
\path[edge] (n0)  -- node[lab]{inst} (n2) ;
\path[edge] (n2) -- node[lab]{running} (n1) ;
\path[edge] (n3)  -- node[lab]{deployed} (n4) ;
\path[edge] (n3)  -- node[lab]{inst} (n5) ;
\path[edge] (n5) -- node[lab]{running} (n4) ;
\userdefinedmacro
\end{tikzpicture}
\renewcommand{\userdefinedmacro}{\relax}

%% file: figures/ECUs-export/deploy.tikz
%
\begin{tikzpicture}[
scale=\tikzscale*1.2]
\node[node] (n0)  at (0.800, -0.200) {\ml{\textbf{:Cmpnt}}};
\node[node] (n1)  at (1.200, -1.000) {\ml{\textbf{:Node}}};
\path[nacedge](n0) -- node[naclab]{deployed} (n1) ;
\path[newedge] (n0) .. controls (0.300, -0.600) and (0.400, -1.100) ..  (n1) ;
\node[newlab] at (0.400, -0.850){\mlc{<<++>>\\deployed}};
\userdefinedmacro
\end{tikzpicture}
\renewcommand{\userdefinedmacro}{\relax}

%% file: figures/ECUs-export/createInst.tikz
%
\begin{tikzpicture}[
scale=\tikzscale*1.2]
\node[node] (n0)  at (1.200, -0.100) {\ml{\textbf{:Cmpnt}}};
\node[node] (n1)  at (0.600, -0.900) {\ml{\textbf{:Node}}};
\node[newnode] (n2)  at (1.800, -0.900) {\mlc{<<++>>\\\textbf{:CInst}}};
\node[nacnode] (n3)  at (2.400, -0.100) {\ml{\textbf{:CInst}}};
\path[edge] (n0)  -- node[lab]{deployed} (n1) ;
\path[newedge](n0) -- node[newlab]{\mlc{<<++>>\\inst}} (n2) ;
\path[newedge](n2) -- node[newlab]{\mlc{<<++>>\\running}} (n1) ;
\path[nacedge] (n0)  -- node[naclab]{inst} (n3) ;
\path[nacedge] (n1) .. controls (0.300, -1.300) and (0.350, -1.400) .. (0.350, -1.400).. controls (0.400, -1.550) and (0.800, -1.550) .. (0.850, -1.400).. controls (0.850, -1.400) and (0.900, -1.300) ..  (n1) ;
\node[naclab] at (0.600, -1.450){down};
\userdefinedmacro
\end{tikzpicture}
\renewcommand{\userdefinedmacro}{\relax}

%% file: figures/ECUs-export/destroyInst.tikz
%
\begin{tikzpicture}[
scale=\tikzscale*1.2]
\node[node] (n0)  at (1.200, -0.100) {\ml{\textbf{:Cmpnt}}};
\node[node] (n1)  at (0.600, -0.900) {\ml{\textbf{:Node}}};
\node[delnode] (n2)  at (1.800, -0.900) {\mlc{<<-\,->>\\\textbf{:CInst}}};
\path[deledge] (n0)  -- node[dellab]{\mlc{<<-\,->>\\inst}} (n2) ;
\path[deledge](n2) -- node[dellab]{\mlc{<<-\,->>\\running}} (n1) ;
\userdefinedmacro
\end{tikzpicture}
\renewcommand{\userdefinedmacro}{\relax}

%% file: figures/ECUs-export/shutdown.tikz
%
\begin{tikzpicture}[
scale=\tikzscale*1.2]
\node[node] (n0)  at (0.800, -0.200) {\ml{\textbf{:Node}}};
\node[nacnode] (n1)  at (2.000, -0.200) {\ml{\textbf{:CInst}}};
\path[nacedge](n1) -- node[naclab]{running} (n0) ;
\path[newedge] (n0) .. controls (0.500, -0.600) and (0.550, -0.700) .. (0.550, -0.700).. controls (0.600, -0.850) and (1.000, -0.850) .. (1.050, -0.700).. controls (1.050, -0.700) and (1.100, -0.600) ..  (n0) ;
\node[newlab] at (0.800, -0.700){\mlc{<<++>>\\down}};
\userdefinedmacro
\end{tikzpicture}
\renewcommand{\userdefinedmacro}{\relax}

%% file: figures/ECUs-export/goal.tikz
%
\begin{tikzpicture}[
scale=\tikzscale*1.2]
\node[node] (n0)  at (0.800, -0.200) {\ml{\textbf{n1:Node}}};
\node[node] (n1)  at (2.000, -0.100) {\ml{\textbf{c1:Cmpnt}}};
\node[node] (n2)  at (2.000, -0.900) {\ml{\textbf{:CInst}}};
\node[node] (n3)  at (3.000, -0.100) {\ml{\textbf{c2:Cmpnt}}};
\node[node] (n4)  at (3.000, -0.900) {\ml{\textbf{:CInst}}};
\path[edge] (n0) .. controls (0.500, -0.600) and (0.550, -0.700) .. (0.550, -0.700).. controls (0.600, -0.850) and (1.000, -0.850) .. (1.050, -0.700).. controls (1.050, -0.700) and (1.100, -0.600) ..  (n0) ;
\node[lab] at (0.800, -0.750){down};
\path[edge](n1) -- node[lab]{inst} (n2) ;
\path[edge](n3) -- node[lab]{inst} (n4) ;
\path[edge](n2) -- node[naclab]{running} (n0) ;
\userdefinedmacro
\end{tikzpicture}
\renewcommand{\userdefinedmacro}{\relax}

%% file: 4_evaluation.tex
\section{Evaluation}
\label{sec:evaluation}

\paragraph{Heuristics}

We compared our heuristic function ($h_{abs}$) against a similarity-based heuristic which resembles heuristic functions employed in related work~\cite{edelkamp06_heuristicSearch,snippe11_usingHeuristicSearch}.
Both heuristics have been implemented in GROOVE~\cite{kastenberg06_groove}, a tool for state space generation and verification of graph grammars, to conduct the experiments.

The similarity-based heuristic ($h_{sim}$) counts the number of nodes and edges that exist in both the current configuration and the target configuration.
It relies on the types of nodes and edges to judge whether a node or edge is counted as existing.
More precisely, it puts the type of each node and edge of a configuration into a multiset and takes the cardinality of the intersection of the current configuration's multiset and the target configuration's multiset as a similarity measure.
The heuristic value is then defined as the additive inverse of this measure.

\paragraph{Search algorithms}

Both heuristic functions are evaluated in combination with greedy best-first and a variant of enforced hill-climbing.

Greedy best-first (GBF) is a well-known search algorithm for informed search.
It uses a closed list and an open list of states.
After expanding a state, i.e., all successor states have been generated, the state is placed in the closed list.
For each new successor state found, its heuristic value is computed and it is placed into the open list.
The decision which state to expand next is based on the heuristic values of the states in the open list.

Enforced hill-climbing (EHC) is a local search algorithm.
In each iteration it performs a breadth-first search from the current state until it finds a state with a better heuristic value.
When such a state is found, it updates the current state and continues with the next iteration.
We use a modified EHC that applies greedy best-first search instead of breadth-first search in each iteration.

\paragraph{Problem domains}

We used two problem domains for our experiments: Blocks World and ECUs.

Blocks World is a classical problem domain in the area of AI planning.
It constitutes of a table with a set of cubes that can be stacked upon each other.
A cube can only be moved if there are no other cubes on top of it and there is only one arm that can hold a cube, i.e., two cubes cannot be moved at the same time.
Finding an optimal solution in this domain has been shown to be NP-hard \cite{gupta92_blocksWorld}.

The ECUs domain functions as explained in Section~\ref{sec:approach}.
In contrast to the Blocks World domain which does not involve the instantiation of objects, the ECUs domain contains rules creating new objects.

\paragraph{Experiment setup}

For the Blocks World domain, we used 8 different problem sizes (4, 6, 8, 10, 12, 14, 16, and 18 blocks) and 4 different problem instances (cross product of 2 random initial and target configurations) per problem size.

For the ECUs domain, we used 4 different problem sizes (2, 3, 4, and 5 ECUs), each with 4 different problem instances.
Two of these problem instances had the same number of component instances running in the initial configuration as ECUs were available.
The other two problem instances used an additional component instance.
Each target configuration specified every second ECU (rounding down at odd numbers of ECUs) to be shut down.

The experiments were conducted on a Dual Intel Xeon E5520 compute server with 16 (virtual) cores running at 2.27GHz.
Each experiment was given 4 cores and 4GB of RAM.
If no plan could be computed within 20 minutes, the job was terminated.

\paragraph{Results}

\begin{figure}[!t]
  \centering
  \includegraphics[width=.8\textwidth]{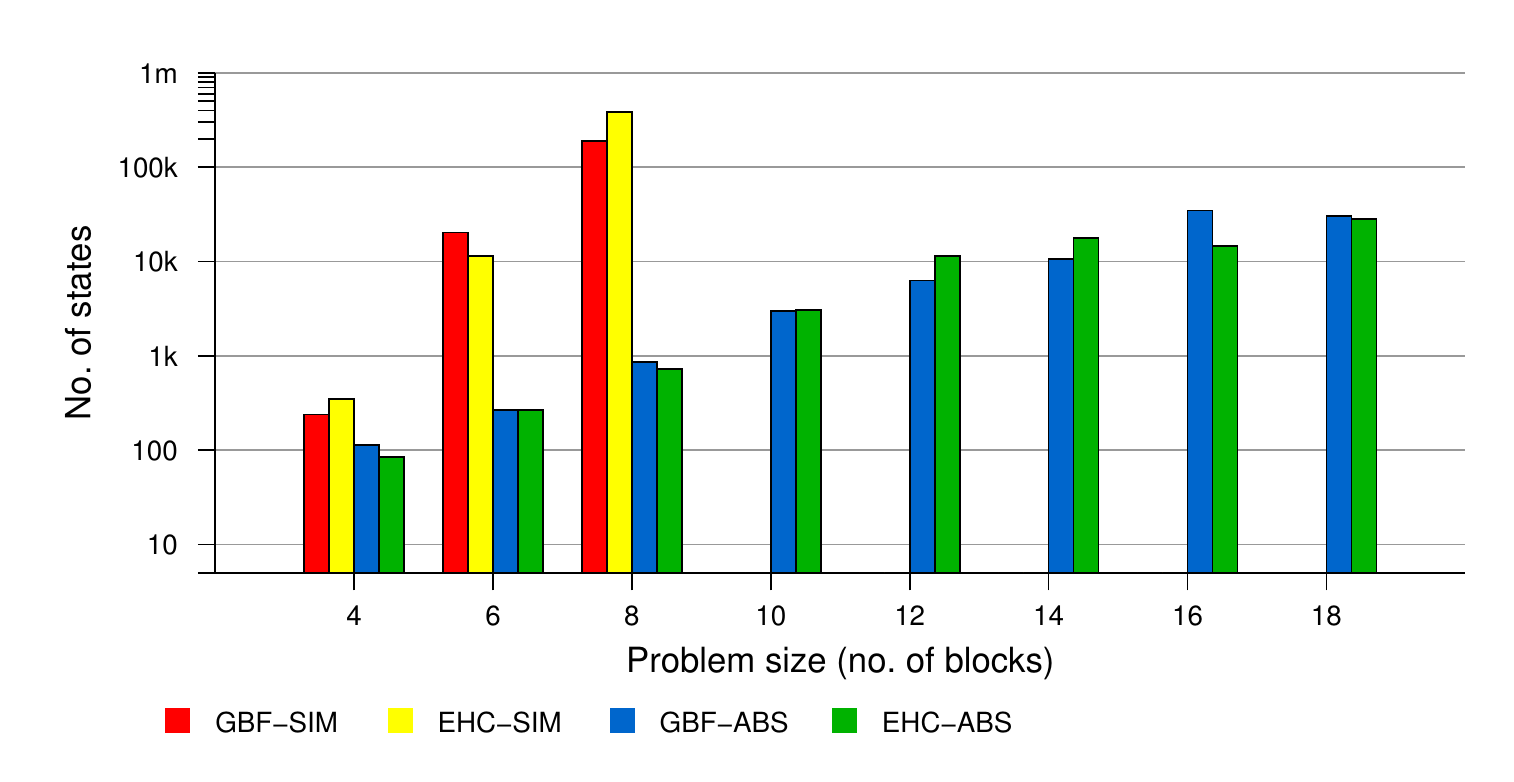}
  \caption{Histogram of the average number of states in Blocks World domains}
  \label{fig:ploticus-BlocksWorld-numberOfStates}
\end{figure}

\begin{figure}[!t]
  \centering
  \includegraphics[width=.8\textwidth]{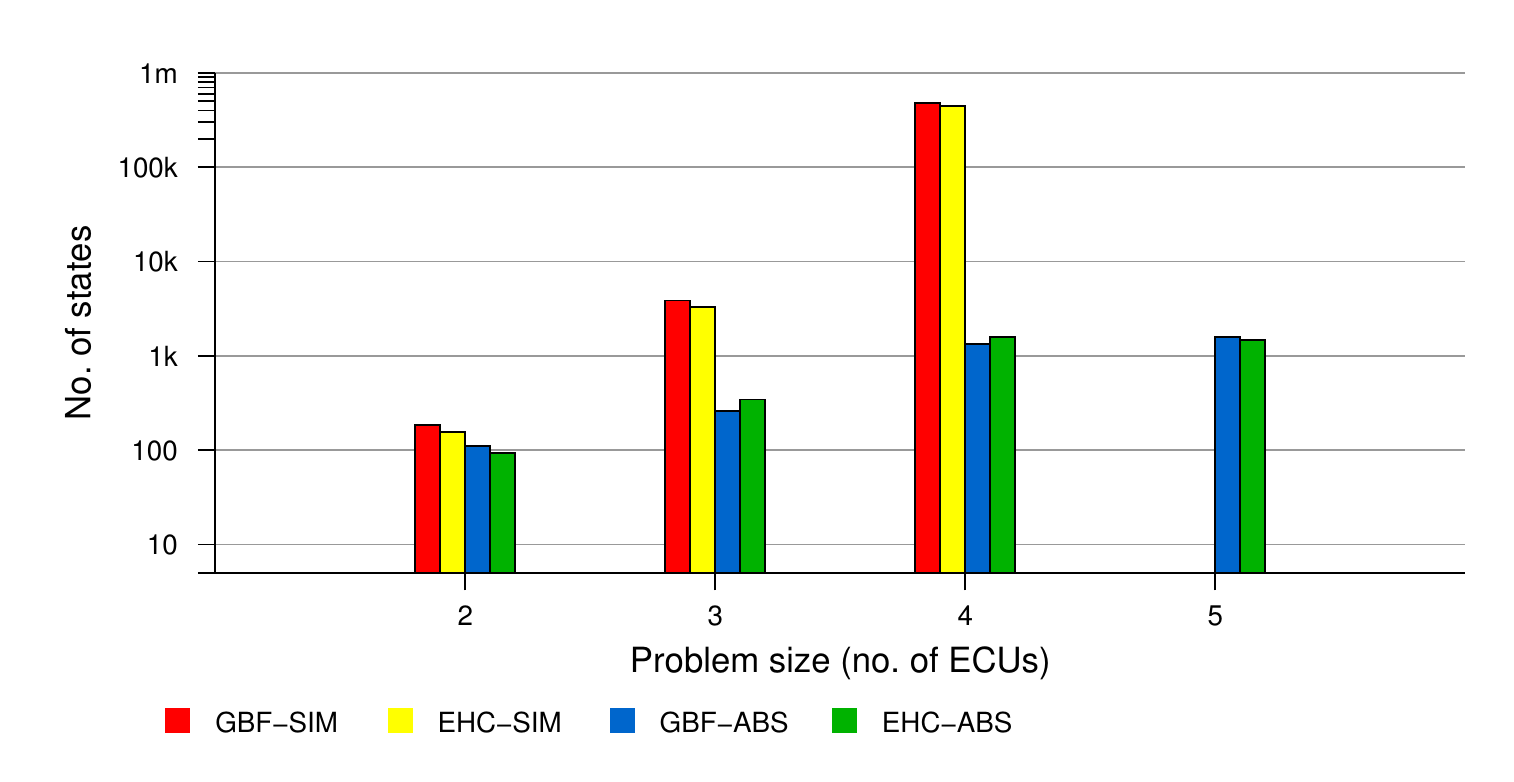}
  \caption{Histogram of the average number of states in ECUs domains}
  \label{fig:ploticus-ECUs-numberOfStates}
\end{figure}

First, we give an overview of the number of generated states for each combination of heuristic function and search algorithm.
The number of generated states means the number of states that have been found during the generation of the concrete planning problem's state space.
This number is a measure for how well the employed heuristic prunes the state space.
It does not include abstract states generated by $h_{abs}$.
Figure~\ref{fig:ploticus-BlocksWorld-numberOfStates} shows a histogram of the average number of states for the BlocksWorld domain, Figure~\ref{fig:ploticus-ECUs-numberOfStates} for the ECUs domain.
Note the logarithmic scale in both histograms.
With increasing problem size $h_{abs}$ makes its superiority clear.
Combinations with $h_{sim}$ failed to provide a solution within 20 minutes for the problems of size 10 blocks and above (on the BlocksWorld domain) and 5 ECUs (on the ECUs domain).
There is no significant difference in performance between GBF and EHC.

\begin{figure}[!t]
  \centering
  \includegraphics[width=.8\textwidth]{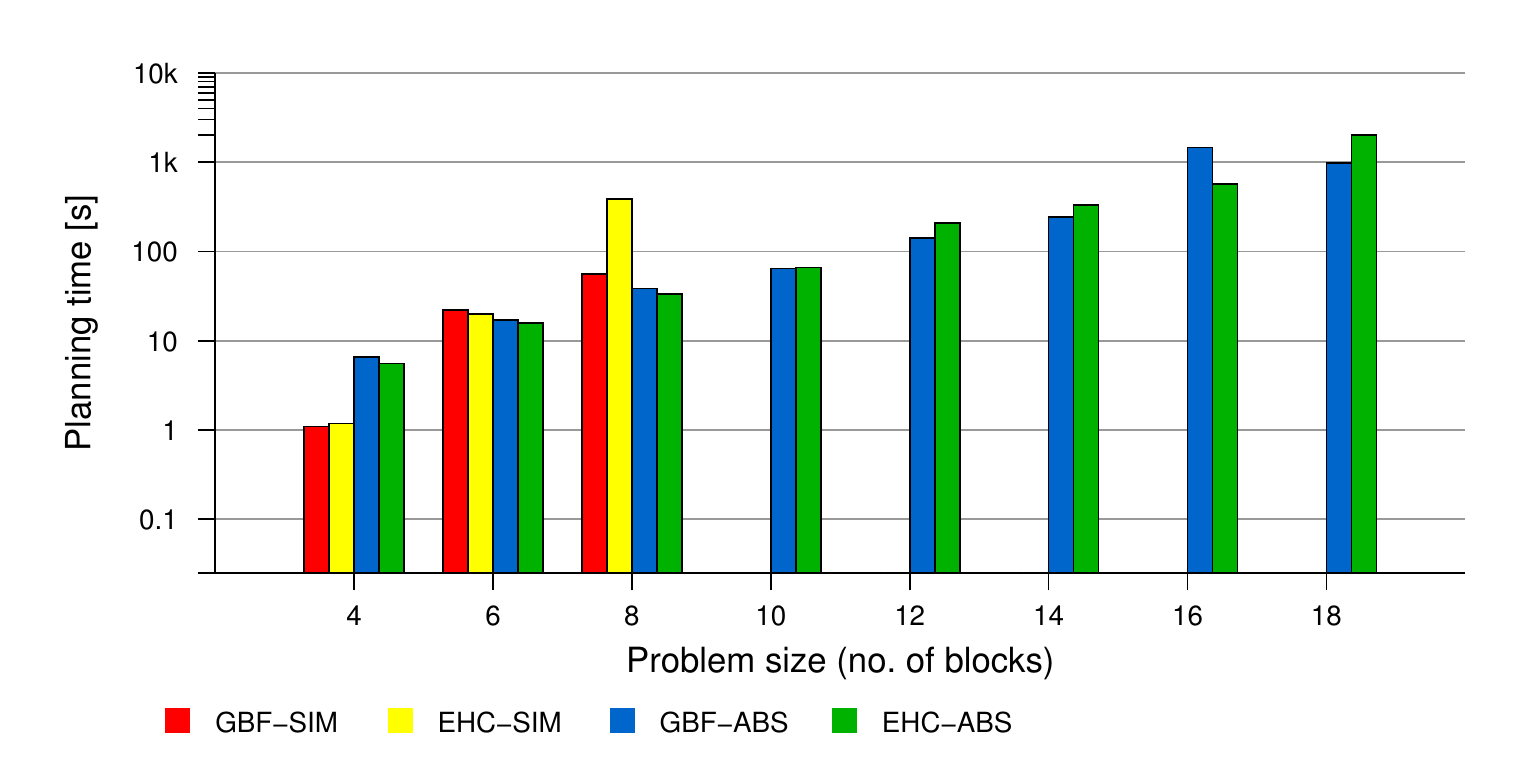}
  \caption{Histogram of the average planning times in Blocks World domains}
  \label{fig:ploticus-BlocksWorld-totalTime}
\end{figure}

\begin{figure}[!t]
  \centering
  \includegraphics[width=.8\textwidth]{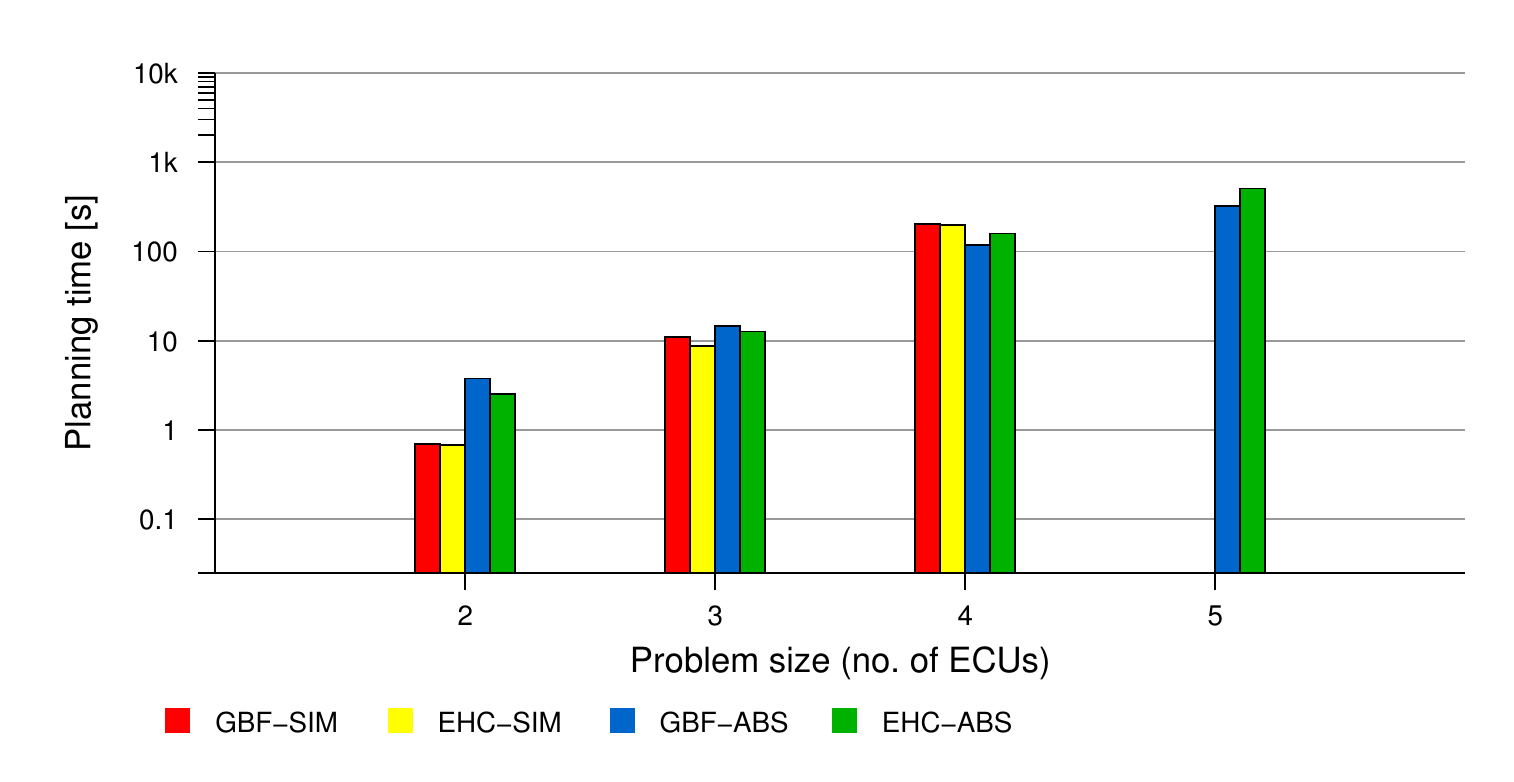}
  \caption{Histogram of the average planning times in ECUs domains}
  \label{fig:ploticus-ECUs-totalTime}
\end{figure}

Considering the average planning times in Figures~\ref{fig:ploticus-BlocksWorld-totalTime} and \ref{fig:ploticus-ECUs-totalTime}, we can observe that $h_{sim}$ performs better than $h_{abs}$ on small domains.
The performance of $h_{abs}$ on small domains is worse than that of $h_{sim}$ because the computation cost of finding a relaxed plan is in general much higher than the computation costs of counting the number of nodes and edges in a state.
While $h_{sim}$ consumes only approx.\ 4\% of the planning time, $h_{abs}$ consumes over 89\% of the planning time (independently of whether used by GBF or EHC).
The performance changes to the favor of $h_{abs}$ as the problem size increases.
The planning time of $h_{abs}$ scales better than the planning time of $h_{sim}$, which is expected since the number of generated states also scales better.
Note that we can only make an appropriate comparison of the scaling behavior, not the \emph{absolute} planning times, because our implementation is not optimized for efficiency.

%% file: 5_related_work.tex
\section{Related Work and Further Discussion}
\label{sec:related_work}

Our heuristic function is mainly inspired by the planning system Fast-Forward (FF)~\cite{hoffmann01_ffJAIR}.
FF is a forward-chaining planner with a heuristic function that uses the solution length of a relaxed problem as one of the main features.
It won the 2nd International Planning Competition (IPC-2000), which led to a shift of planning research towards heuristic-guided approaches.
Variants of its techniques are used in many of today's state-of-the-art planners, like SGPlan~\cite{chen06_sgplan} or LAMA~\cite{richter10_lama}.

In contrast to our system, which uses label propagation to find the relaxed plan, FF computes the relaxed plan by a backward search on a structure called the \emph{planning graph}~\cite{blum97_planningGraph}.
Such a planning graph roughly resembles our list of successor graphs in the abstraction.
By applying this idea to graph transformation systems instead of PDDL's propositional state representations, we face two important differences, especially with the combination of dynamic object creation and the support for NACs.
\begin{itemize}
  \item Dynamic object creation can lead to an explosion of the graph size during the creation of successor graphs in the abstraction.
  This is not an issue in PDDL-based planners because they do not support dynamic object creation.
  \item
  The equivalent to NACs in PDDL are negative existential quantifications over conjunctive facts.
  They are usually solved by compiling them away, i.e., translating them into DNF, which results in a blowup of the propositional domain representation, cf.~\cite{hoffmann01_ffJAIR}.
  In our approach, we avoid such a blowup by building the support for NACs directly into the abstract planning algorithm.
\end{itemize}

Another approach that directly plans on graph transformation system is \cite{estler11_heuristic}.
They also use search algorithms like A* or Best First to search through the state space, but they employ a \emph{domain-specific} heuristic.
An obvious disadvantage is that a heuristic suitable for the given application domain has to be developed first.
To overcome the disadvantage, they use an approach to learn heuristic functions automatically.
A learning algorithm derives a \emph{regression function} that predicts the costs of solving the problem from a given state.
To derive the regression function, the learning algorithm needs a predefined declaration of state features and a training set with problem instances.
While this solution is an improvement from developing heuristic functions manually, it still requires the developer to declare a set of state features that is suitable for the given application domain.

In the introduction, we mentioned approaches that translate the GT planning problem into PDDL as an alternative to planning directly on graph transformation systems.
This has been done in \cite{tichy11_planningSelfAdaptation} for the story pattern formalism and in \cite{ziegert13_temporalReconfigurationPlans}, an extension of the former by a transactional concept to support durations that are specified for reconfigurations.
In both of them, the straightforward solution to support NACs is to translate them into negative existential quantifications over all objects of the forbidden objects' types.
The major disadvantage of these approaches is the restricted expressiveness of PDDL.
Because PDDL does not support the dynamic creation of objects, these approaches use a modeling workaround, which requires a fixed maximal number of objects per type.

%% file: 6_conclusion.tex
\section{Conclusion and Future Work}
\label{sec:conclusion}

In this paper, we presented an approach to planning with graph transformations.
It features a domain-independent heuristic function that uses the solution of an abstraction of the problem as guidance.
We further showed that it performed better than a similarity-based heuristic function which resembles heuristic functions employed in related work~\cite{edelkamp06_heuristicSearch,snippe11_usingHeuristicSearch}.

In future work we plan to further improve our heuristic planning approach.
One idea is to adapt the notion of \emph{helpful actions}, cf.\ \cite{hoffmann01_ffJAIR}, to graph transformation planning.
An action is called helpful in the concrete planning problem, if it is applied in the first step of the parallel relaxed plan.
By considering only helpful actions when expanding a state, we expect to achieve a strong performance improvement.

When we finished developing a GTS-based planner that is satisfyingly performant, we intend to do a detailed evaluation comparing its efficiency to the efficiency of running off-the-shelf planning systems on PDDL models that have been generated out of the GTS by a translator.
We are specifically interested in finding out whether one approach dominates the other one in general or whether this depends on properties of the application domain.